\documentclass[conference]{IEEEtran}
\IEEEoverridecommandlockouts
\usepackage[top=1.9cm, bottom=2.54cm, left=1.6cm, right=1.6cm]{geometry}
\usepackage{cite}
\usepackage{amsmath,amssymb,amsfonts}
\usepackage{algorithmic}
\usepackage{graphicx}
\usepackage{textcomp}
\usepackage{xcolor}
\usepackage{url}
\def\BibTeX{{\rm B\kern-.05em{\sc i\kern-.025em b}\kern-.08em
    T\kern-.1667em\lower.7ex\hbox{E}\kern-.125emX}}
\newcommand{\linebreakand}{}
\begin{document}

\title{LLM-Cave: A benchmark and light environment for large language models reasoning and decision-making system\\
\thanks{This work was supported in part by the National Natural Science Foundation of China under Grant 624B2074 and Grant 62233011.}
}

\author{\IEEEauthorblockN{1\textsuperscript{st} Huanyu Li}
\IEEEauthorblockA{\textit{College of Artificial Intelligence} \\
\textit{Nankai University}\\
Tianjin, China \\
2211286@mail.nankai.edu.cn}
~\\
\and
\IEEEauthorblockN{2\textsuperscript{nd} Zongyuan Li}
\IEEEauthorblockA{\textit{College of Artificial Intelligence} \\
\textit{Nankai University}\\
Tianjin, China \\
2120230524@mail.nankai.edu.cn}
~\\
\linebreakand
\IEEEauthorblockN{4\textsuperscript{th} Xian Guo*}
\IEEEauthorblockA{\textit{College of Artificial Intelligence} \\
\textit{Nankai University}\\
Tianjin, China \\
guoxian@nankai.edu.cn \\
*Corresponding Author}
~\\
\and
\IEEEauthorblockN{3\textsuperscript{rd} Wei Huang}
\IEEEauthorblockA{\textit{College of Artificial Intelligence} \\
\textit{Nankai University}\\
Tianjin, China \\
huangw@mail.nankai.edu.cn}

}

\maketitle

\begin{abstract}
Large language models (LLMs) such as ChatGPT o1, ChatGPT o3, and DeepSeek R1 have shown great potential in solving difficult problems. However, current LLM evaluation benchmarks are limited to one-step interactions. Some of the existing sequence decision-making environments, such as TextStarCraftII and LLM-PySC2, are too complicated and require hours of interaction to complete a game. In this paper, we introduce LLM-Cave, a benchmark and light environment for LLM reasoning and decision-making systems. This environment is a classic instance in the era of Symbolism. Artificial intelligence enables the agent to explore the environment and avoid potential losses by reasoning about nearby dangers using partial observable state information. In the experiment, we evaluated the sequential reasoning ability, decision-making performance and computational efficiency of mainstream large language models (LLMs) such as GPT-4o-mini, o1-mini, and DeepSeek-R1. Experiments show that while Deepseek-R1 achieved the highest success rate on complex reasoning tasks, smaller models like 4o-mini significantly narrowed the performance gap on challenges by employing Chain of Speculation and Planner-Critic strategies, at the expense of reduced computational efficiency. This indicates that structured, multi-step reasoning combined with an LLM-based feedback mechanism can substantially enhance an LLM's decision-making capabilities, providing a promising direction for improving reasoning in weaker models and suggesting a new reasoning-centered benchmark for LLM assessment. Our code is open-sourced in \url{https://github.com/puleya1277/CaveEnv}.

\end{abstract}

\begin{IEEEkeywords}
Large language models, reasoning, benchmark, sequential decision-making
\end{IEEEkeywords}

\section{Introduction}

In recent years, unprecedented advancements have been witnessed in large language models, exemplified by models such as GPT, Claude, Grok, Gemini and Deepseek. These models, trained on vast corpora of text and leveraging scalable transformer-based frameworks, have achieved remarkable performance across diverse natural language processing tasks, including text generation, translation, and question-answering. Beyond these capabilities, their emergent abilities in multimodal understanding, code generation, and context-aware interaction have positioned LLMs as foundational tools in artificial intelligence research. The integration of techniques like reinforcement learning from human feedback (RLHF) and reinforcement learning based finetune (RLFT) have further enhanced their alignment with human intent or task requirements, enabling deployment in real-world applications ranging from education to creative industries.

A critical frontier in LLM development lies in improving their reasoning abilities. Techniques such as chain-of-thought prompting \cite{CoT}, LLM self-reflection \cite{2025arXiv250213388}, Retrieval-Augmented Generation and online deep-search have enabled models to tackle complex logical, mathematical, and inferential tasks. These advancements have catalyzed societal transformations, such as AI gaming \cite{hua2024gametheoreticllmagentworkflow}, AI-assisted financial decision-making \cite{xiao2024tradingagents}, robot navigation \cite{shah2023lmnav} \cite{doma2024llm} \cite{dorbala2023embodied}, and manipulation \cite{jin2024robotgpt} \cite{liu2024enhancing}. However, the reasoning ability of current models is not exactly equal to logic ability. Current QA benchmarks may already exist in the pre-training or fine-tuning process, which raises concerns about the reliability of the actual logic abilities, especially in environments that are not trained in the pre-training process.

Current benchmarks for evaluating the reasoning ability of LLMs, such as MMLU, GSM8K, and Humanity's Last Exam, predominantly focus on single-step question-answering (QA) tasks. While effective for measuring factual recall or short deductive reasoning, these datasets fail to capture the complexity of real-world problems requiring multi-step planning, iterative refinement, and dynamic environment interaction. Such limitations hinder the assessment of the LLMs' ability to manage long-term tasks, where errors compound over sequential decisions,  and adaptive strategies are essential for success. Consequently, a gap exists between existing benchmarks and the demands of practical, multi-step decision-making scenarios.

For long-term decision-making, recent efforts have introduced sequential decision-making environments such as TextStarCraft2 \cite{2023arXiv231211865M} and LLM-PySC2 \cite{LLM-PySC2}, which provide interfaces for the strategic game StarCraft II through text-based interfaces. While these platforms enable research into multi-step reasoning and real-time planning, their reliance on heavy computational infrastructure, complex rule systems, and extensive API dependencies creates significant overhead. Such burdens limit accessibility for researchers, hinder rapid experimentation, and obscure the interpretability of model behavior—issues that demand lightweight, scalable alternatives for widespread adoption.

\begin{figure}[htbp]
  \centering
  \includegraphics[width=0.48\textwidth]{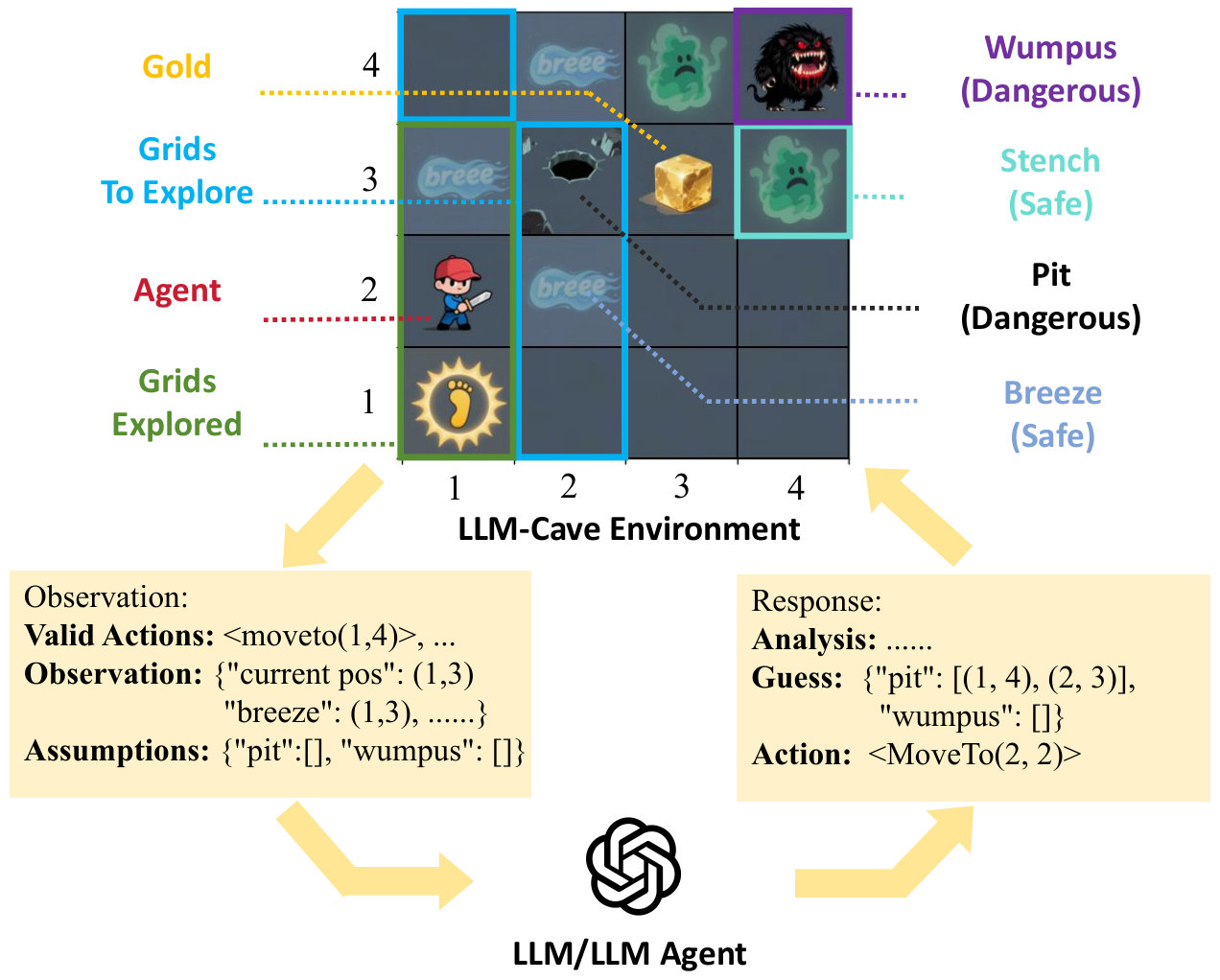}
  \caption{\textbf{LLM-Cave Environment.} 
  In the LLM-Cave environment, the LLM controls a game agent; the agent has to explore the cave and find the gold. In the cave, the agent can only get the information of the current grid, while there are pits (holes) and wumpus (monster) that may kill the agent. Near the pits and wumpus are breeze and stench. The LLM should reason the position of the pit and wumpus according to the observed information of breeze and stench to safely explore the cave and find the gold.  }
  \label{fig1}
\end{figure}

In this paper, we introduce LLM-Cave, a novel benchmark framework and lightweight environment specifically designed for evaluating and enhancing the reasoning and decision-making abilities of Large Language Models (LLMs). This environment not only tests the ability of large language models in planning and reasoning but also their flexibility in the face of complex problems. The environment maintains a lightweight design, facilitating ease of deployment and interpretability of results while precisely capturing the intricacies inherent in sequential reasoning. Within LLM-Cave, we further implement the Chain of Speculation and Planner-Critic mechanisms. The performer-critic mechanism gives the model a “self-reflective” capability. Without human intervention, the model can independently evaluate its own decisions and adjust its direction when necessary. This makes the entire decision-making process more robust and reliable.

In our experiments, we evaluated the sequential reasoning capabilities of several prominent LLMs, including OpenAI GPT-4o-mini, o1-mini, Claude 3.5, Gemini 2.5 Flash, and DeepSeek-R1, within the LLM-Cave environment. The experimental results show clear differences. On the one hand, reasoning-oriented models such as DeepSeek-R1 tend to be able to deduce reasonable strategies with little external prompts. On the other hand, models that lack reasoning ability often struggle to make optimal moves when confronted with a problem. However, after employing our Chain of Speculation and Planner-Critic collaborative mechanisms, performance among weaker models improved notably. Compared to their zero-shot or initial baseline attempts, these models exhibited a reduction in logical errors and an increased success rate.

Our contributions include:

(1) We propose a lightweight, text-based environment explicitly designed for assessing the reasoning and decision-making capabilities of LLMs. 

(2) We introduce the Chain of Speculation inference strategy, which requires the LLM to make a guess about the hazard and explicitly propagate the guess to the next step.

(3) We develop an LLM-driven dual-role feedback framework in which a Planner generates decisions informed by the current Chain of Speculation while a Critic evaluates the correctness of these outputs and proposes optimizations. 

(4)We assessed the decision-making ability and computing efficiency of leading large-language models (LLMs).

The code of the LLMCave environment has been open-so/
*-urced in GitHub \url{https://github.com/puleya1277/CaveEnv}.

\section{Related Works}

\subsection{LLM Decision making}

LLM decision-making problems have been studied for several years since ChatGPT came into view.  Early works such as Stanford Town \cite{park2023generative} focused on LLMs’ behaviors, and LLMs exhibit human-like behaviors. After that, more and more works tested LLM's decision-making abilities in games like ChatDev \cite{ChatDev}, CRADLE \cite{tan2024towards}, LlamaRider \cite{MCLlamaRider}, SwarmBrain \cite{shao2024swarmbrainembodiedagentrealtime} and TextStarCraft2. In the field of robotics, LLMs are also used in robotic manipulation and navigation. These works show the potential of LLM in sequential decision-making, but these environments are not widely used as benchmarks for LLMs because of their complexity and high demand for computational resources and token consumption.


\subsection{LLM Reasoning Enhancing Techniques}

In September 2024, OpenAI released the first long CoT model, GPT-o1. Since then, techniques for enhancing LLM reasoning ability have attracted increasing attention. Train-free prompt engineering methods like Chain-of-Draft \cite{CoD} increase LLM's performance directly through input instructions, while Multi-Agent LLM System \cite{tan2024towards} improves the performance by splitting tasks into smaller subtasks, enhancing the tool use and coding abilities  \cite{ToolStar} to better handle problems. For training methods, cold start supervised training, RLHF and RLFT increase the performance by directly updating the parameters of the models.


\section{Preliminaries}

\subsection{Markov Decision Process}

Within the framework of sequential decision modeling, a Markov decision process (MDP) establishes that the state at any given moment is independent of future actions. Over the last decade, MDP has become a standard paradigm for modeling sequential decision-making challenges, including optimal control strategies, policy optimization, reinforcement learning, and collaborative multi-agent systems.

An MDP structure denoted as $MDP(S, A, R,\rho,s_0)$ originating at $s_0$ operates through iterative interactions: At each timestep $t$, the agent observes current environment information $s_t$ of state space $S$ and selects an execution command $a_t$ from action space $A$. The environment subsequently produces an immediate feedback signal $r_{t}$ determined by the reward function $r_{t} = R(s_t, a_t)$, and transitioning to successor state $s_{t+1}$ according to transition dynamics $p_{s_{t+1}} = \rho(s_{t+1}|s_t, a_t)$. For scenarios with incomplete state information, the model extends to partially observable MDP $POMDP(O, S, A, R, \rho, s_0)$, introducing observation domain $O$ where $o_t \in O$ denotes the observation at timestep $t$.



\subsection{LLM Decision Making Process}

For LLM-based decision making, we denote $p$ as profile or system prompt, and the LLM makes decisions according to $$\widehat{a}_t=LLM_p(\widehat{o}_t)$$ where $\widehat{.}$ represents any data in text form. In CoT settings, an LLM will be prompted to respond as: $$ (\widehat{c}_1,\widehat{c}_2,...\widehat{c}_n)_{CoT} = LLM_p(\widehat{I}_t) $$ where content $\widehat{c}_{i+1}$ is generated after $\widehat{c}_i$ and $\widehat{I}_t$ represent input content. 


\section{Method}

The chain of thought (CoT) approach enables the model to explicitly output the analysis process, thereby improving the decision-making capability of large models \cite {CoT}. We build on CoT by introducing two core components to the LLM agent: the Chain of Speculation mechanism for long-term reasoning memory and the Planner-Critic mechanism for decision verification. Below, we describe each component in detail.

\begin{figure}[htbp]
\centering
\includegraphics[width=0.48\textwidth]{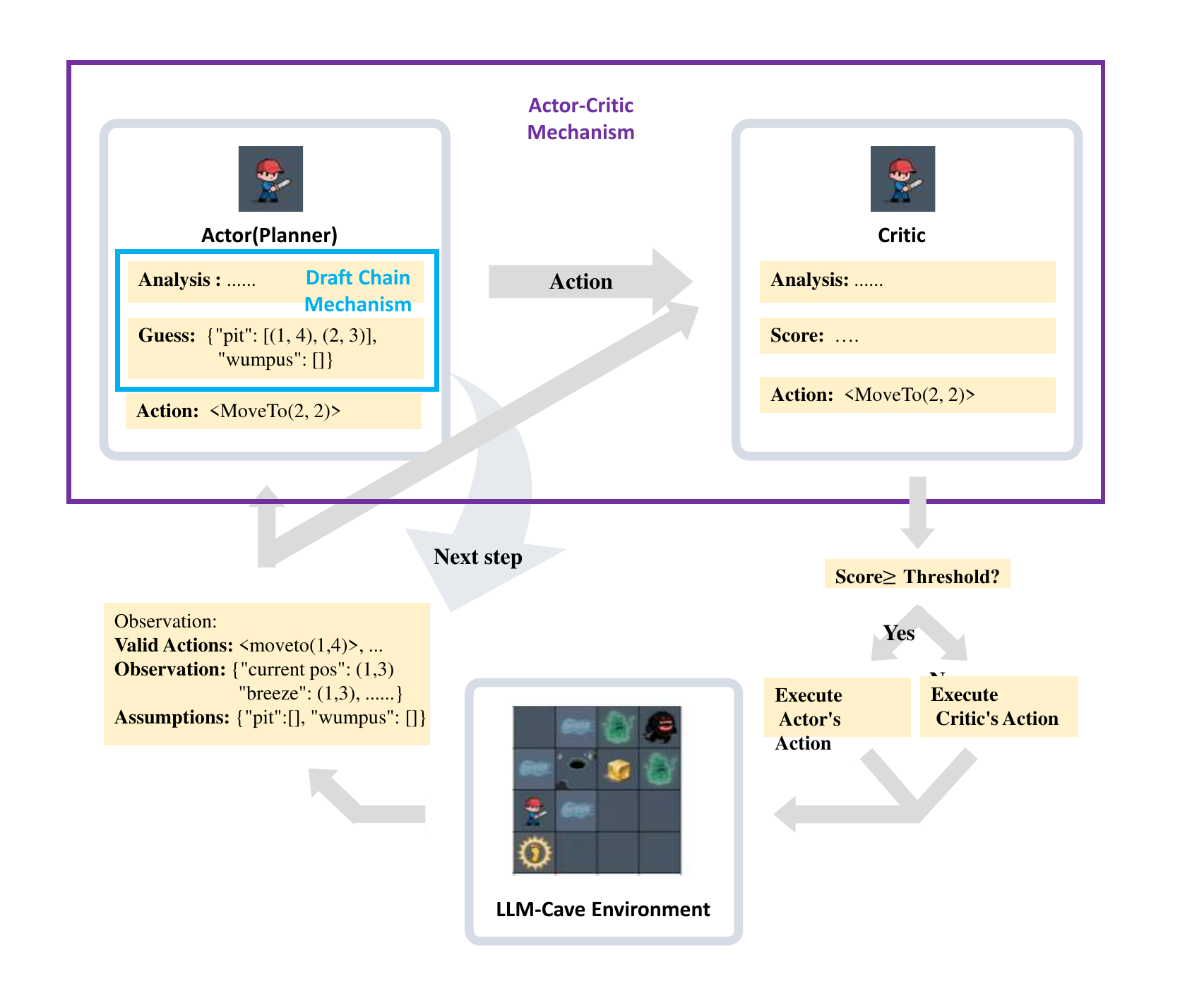}
\caption{\textbf{The workflow of LLM agents interacting with its environment.} 
The Chain of Speculation Mechanism and the Planner-Critic Mechanism are applied within LLM-Cave. The Chain of Speculation maintains explicit hypotheses about pit and Wumpus positions, updating them after each observation. The Planner proposes an action that the Critic scores for safety; actions exceeding a confidence threshold are executed otherwise a safer alternative is supplied. }
\label{fig2}
\end{figure}

\subsection{Chain of Speculation Mechanism}\label{CC}
The Chain of Speculation mechanism aims to provide LLM agents with explicit reasoning memory across multiple interaction rounds. While standard LLM agents implicitly access the dialogue history (including past observations, previous actions, or reasoning) through prompts, they often struggle to maintain a structured and coherent understanding of the environmental state over time. The Chain of Speculation addresses this limitation by enabling the model to explicitly generate and propagate a hypothesis ("draft") regarding the environmental state at each interaction step.

Concretely, the output of the LLM at each interaction is structured into three distinct parts: \textbf{Analysis} (a reasoning process based on current observations), \textbf{Guess} (the model's hypothesis or inference regarding the locations of Wumpus and pits—explicitly formatted in JSON for clarity and ease of subsequent processing), and \textbf{Action} (the decision based on these hypotheses). For instance, upon perceiving a \verb| "stench" | at locations (1,2) and (2,1), the model's Guess might be structured as \verb| "wumpus": [(2,2)]|, and the corresponding Action could be "\verb|<shootright>| to ensure the absence of Wumpus on the right side." This Guess is then appended to the subsequent observations, thus reminding the model of its current inferences when it receives new information. Essentially, the model declares its evolving beliefs explicitly, thereby constructing and maintaining its own knowledge base. 

To clearly illustrate, each interaction round follows the procedure below: 

\begin{itemize}
\item The agent receives the latest observations from the environment (e.g., detecting a breeze or stench at its current position). 
\item These new observations, along with the Guess from the previous round (if available), are incorporated into the prompt. The LLM then outputs a new Analysis, updated Guess, and subsequent Action. The analysis represents the model's reasoning about the immediate environmental cues, Guess encapsulates the agent's updated beliefs about possible Wumpus or pit locations, and Action is chosen based on these updated beliefs. 
\item Newly generated Guesses are stored and passed to the LLM with the next round of observations, thus forming a continuous "draft chain" throughout the task.
\end{itemize}

By iteratively updating its own assumptions, the intelligent body gradually builds up a strong, long-term reasoning ability. In this process, the Chain of Speculation mechanism plays a role similar to that of "internal memory," which serves as the internal state of the large language model. At the same time, this mechanism also has the ability of self-correction - when a new environmental observation conflicts with the current speculation, the model will take the initiative to adjust the original speculation in the following rounds and correct its own cognitive bias step by step. In a sense, this stops the model from "starting from scratch" at every step and allows it to move forward as it accumulates.

\subsection{Planner-Critic Decision Mechanism }
The Planner-Critic decision mechanism enhances decision-making quality by introducing a "second opinion" for each proposed Action. In our framework, we pair the original Planner LLM with a separate Critic LLM, enabling the Critic to evaluate the actions suggested by the Planner. 

Specifically, within this framework, the Planner is responsible for proposing the next Action (leveraging information maintained by the Chain of Speculation as described previously), while the Critic reviews and validates this Action prior to execution. The detailed procedure per round is as follows: 

\begin{itemize}
\item Given the current observation and continually updated Guess, the Planner LLM performs an analysis and outputs its Guess alongside a proposed Action (e.g., "move to position (2,1)" or "pick up the gold"). This represents the agent's initial decision during the current interaction. 
\item The Critic LLM receives relevant context, including the current observation and the Planner's proposed Action. The Critic then evaluates the Action's safety and appropriateness. The Critic responds by assigning a confidence score between 0 and 1 to the Planner's proposed Action and suggests an alternative action it believes would be superior, if any.

\item If the Critic assigns a high confidence score (exceeding a predetermined threshold, such as 0.7) to the Planner's proposed Action, the original Action is executed. Conversely, if the confidence score falls below the threshold, the alternative Action proposed by the Critic is chosen. Subsequently, the environment returns the next observation. 
\end{itemize}

The above Planner-Critic procedure is repeated in every round. The threshold (set as 0.7 in our experiments) acts as a hyperparameter balancing risk and efficiency—lower thresholds indicate the Critic intervenes only when strongly opposing the Planner's choice, whereas higher thresholds prompt the Critic to more frequently override the Planner to ensure safer outcomes.

It is important to note that the Planner and Critic can either use the same underlying LLM with different prompts or two entirely different models. For fairness and comparability, our implementation uses the same model for both Planner and Critic when evaluating a specific LLM. Thus, the Critic essentially functions as an immediate validator. 

The Planner-Critic mechanism significantly reduces the occurrence of overt mistakes. For example, if the Planner chooses an action that could result in falling into a pit due to insufficient reasoning, the Critic—armed with identical observational data but tasked specifically with critical assessment—can identify this danger by interpreting clues such as "breeze." Consequently, the Critic assigns a low confidence score to this hazardous Action, thereby preventing its execution, and instead suggests a safer alternative. Empirically, this process substantially lowers the incidence of catastrophic failures (such as agent deaths) in our experiments. Although the drawback is an increased computational cost (as each step effectively doubles model invocations), the enhanced safety and higher success rates achieved through improved reasoning justify this additional overhead in tasks where correct inference is paramount.

\section{Experiments}

We conducted a series of experiments in LLM-Cave to evaluate the impact of mechanisms such as the Chain of Speculation and Planner-Critic on agent performance. LLM-Cave is a classic grid-world environment that requires an agent to infer the location of hidden dangers from partial information, making it a suitable testbed for evaluating reasoning and decision-making capabilities. The metrics used to measure performance include the average total reward, success rate, and Wumpus kill rate, the average number of steps the agent took to complete the task and the average reward per step. In addition, we introduce computational performance metrics, including average latency, tokens, average cost, and tokens per second. Each experimental configuration was run for 150 trials(3×3 board: zero pits one wumpus, one pit zero wumpuses, one pit one wumpus; 4×4 board: one pit one wumpus, two pit one wumpus, three pit one wumpus, with 25 trials per condition, each using a different random seed) to ensure reliable statistics.

\subsection{LLM-Cave}

We have developed a text-based Wumpus World simulation environment tailored for LLM-agent interaction. The environment consists of an $n \times n$ grid of rooms, consistent with the classical definition of the Wumpus problem. Within this grid, a single dangerous Wumpus and a number of bottomless pits (ranging from 0 to 3 in our setup) are hidden, along with a golden treasure placed in one of the rooms. The agent always starts in the bottom-left room $(1,1)$, which, along with its two adjacent rooms, is guaranteed safe. The primary objective for the agent is to locate the gold while avoiding death.

\begin{figure}[htbp]
\centering
\includegraphics[width=0.48\textwidth]{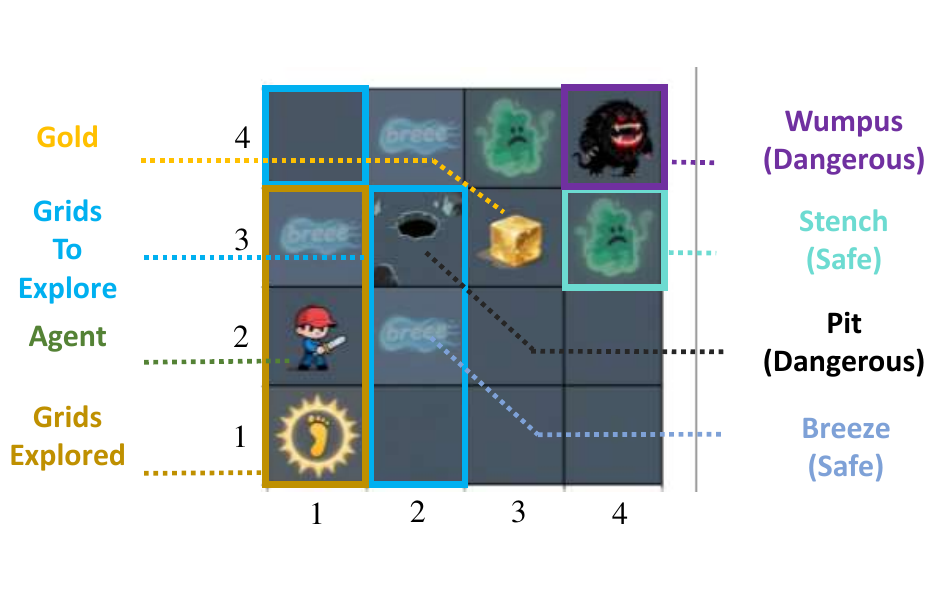}
\caption{\textbf{visualization of LLM-Cave.}}
\label{fig}
\end{figure}

To simplify decision-making, we allow the agent to move by explicitly specifying the coordinates of an adjacent, unexplored room (i.e., a cell adjacent to any previously explored room) while remaining within grid boundaries. When the agent enters a room containing the gold, it automatically collects the treasure. Additionally, the agent possesses a single arrow, which can be shot in a chosen direction; if the Wumpus is present along that trajectory, it is killed (indicated by a scream), and the agent earns bonus points. The task terminates either when the agent successfully collects the gold (success) or when it falls into a pit or is killed by the Wumpus (failure). Notably, we do not require the agent to explicitly exit the cave after collecting the gold—retrieving the treasure itself is considered a success. However, since the gold may be unreachable due to randomized layouts, the agent is permitted to perform an "exit" action, which terminates the episode without incurring a reward or penalty.

At each step, the environment provides textual perceptual feedback describing the agent's current sensory experience. We implement standard Wumpus World signals: if any adjacent (up, down, left, or right) cell contains a pit, a \textit{breeze} is perceived; if the Wumpus is nearby, the agent detects a \textit{stench}; and if the gold is present in the current room, the agent perceives a \textit{glitter}. We aggregate all prior perceptual feedback into a structured \texttt{JSON} observation comprising eight fields:

\begin{itemize}
    \item \texttt{'number of Wumpus'}  
    \item \texttt{'number of pit'}
    \item \texttt{'Current Position'}
    
    \item \texttt{'No breeze or stench is detected in these locations'}
    \item \texttt{'A breeze is detected in the following locations'}
    \item \texttt{'A stench is detected in the following locations'}
    \item \texttt{'When you have confirmed that the corresponding locations are safe, prioritize exploring these areas'}: suggestions for unexplored but adjacent rooms and potential shooting actions (if the agent still has an arrow)
    \item \texttt{'The situation with the arrows'}: including whether the arrow was fired, in what direction, and whether a scream was heard
\end{itemize}

The agent must analyze this observation to infer the likely locations of danger and the treasure. Crucially, the agent never directly observes the precise positions of the pits or the Wumpus. Instead, it must deduce such information from indirect perceptual signals, requiring logical reasoning to avoid hazards.

\begin{figure}[htbp]
  \centering
  \includegraphics[width=0.48\textwidth]{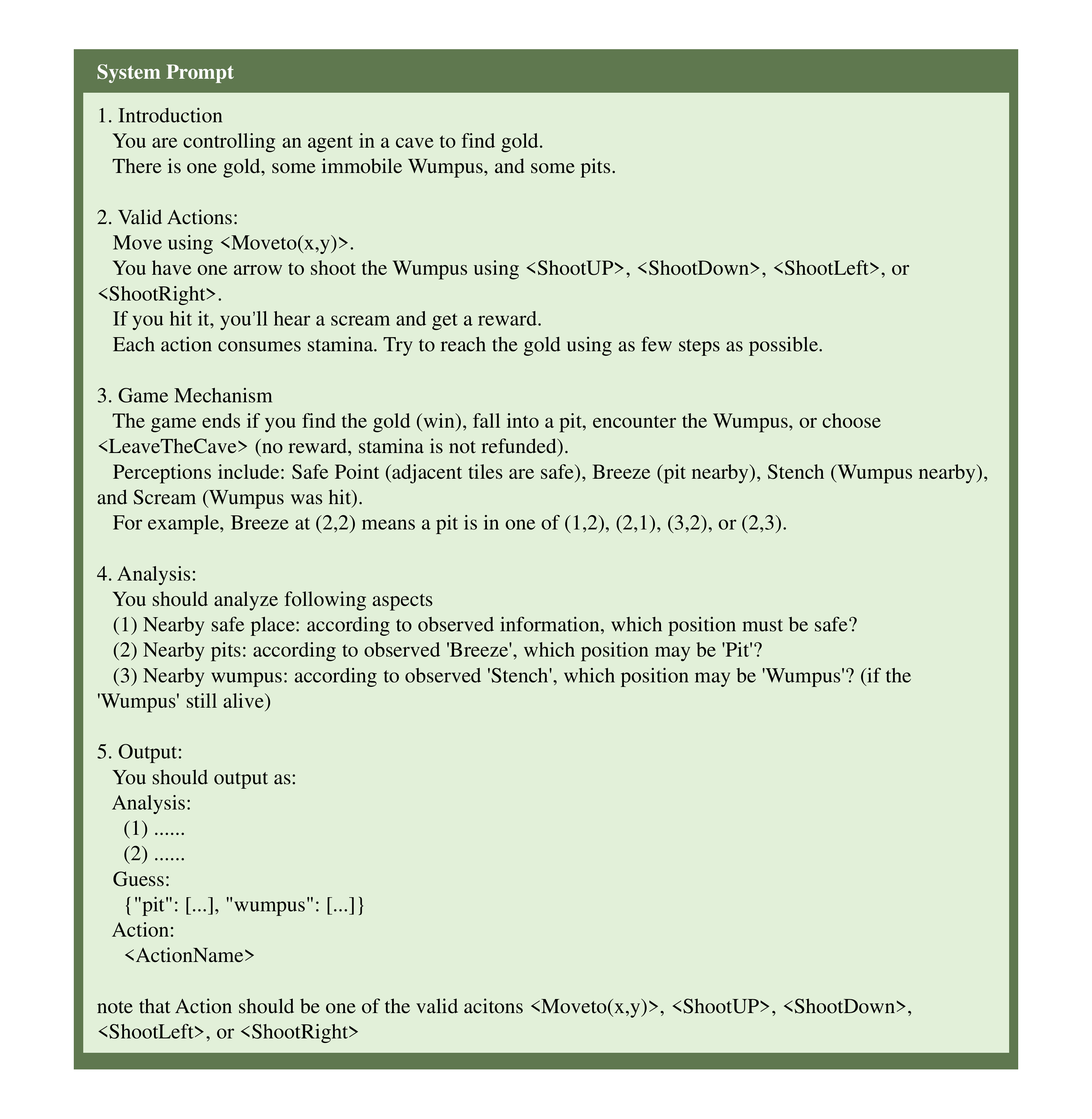}
  \caption{\textbf{System prompt for LLMs.} 
  }
  \label{fig3}
\end{figure}

To quantify agent performance, we adopt a reward structure inspired by the traditional Wumpus World. Each episode begins with a base score of 50 points. Every action (i.e., movement) results in a minor penalty of $-1$ point, encouraging efficiency. This penalty remains minor unless the agent performs many redundant steps. Successfully acquiring the gold yields a bonus of $+50$ points. Fatal outcomes—falling into a pit or being killed by the Wumpus—result in penalties of $-20$ or $-30$ points, respectively. Successfully killing the Wumpus with the arrow earns $+20$ bonus points.

Each episode terminates upon success (gold collection), failure (death), or exceeding the step limit (50 steps per episode to prevent infinite loops). Thus, an ideal episode may achieve a normalized score slightly above 100(if the Wumpus is killed), while poor performance with extensive movement and eventual death can result in negative scores.

\begin{figure}[htbp]
\centering
\includegraphics[width=0.48\textwidth]{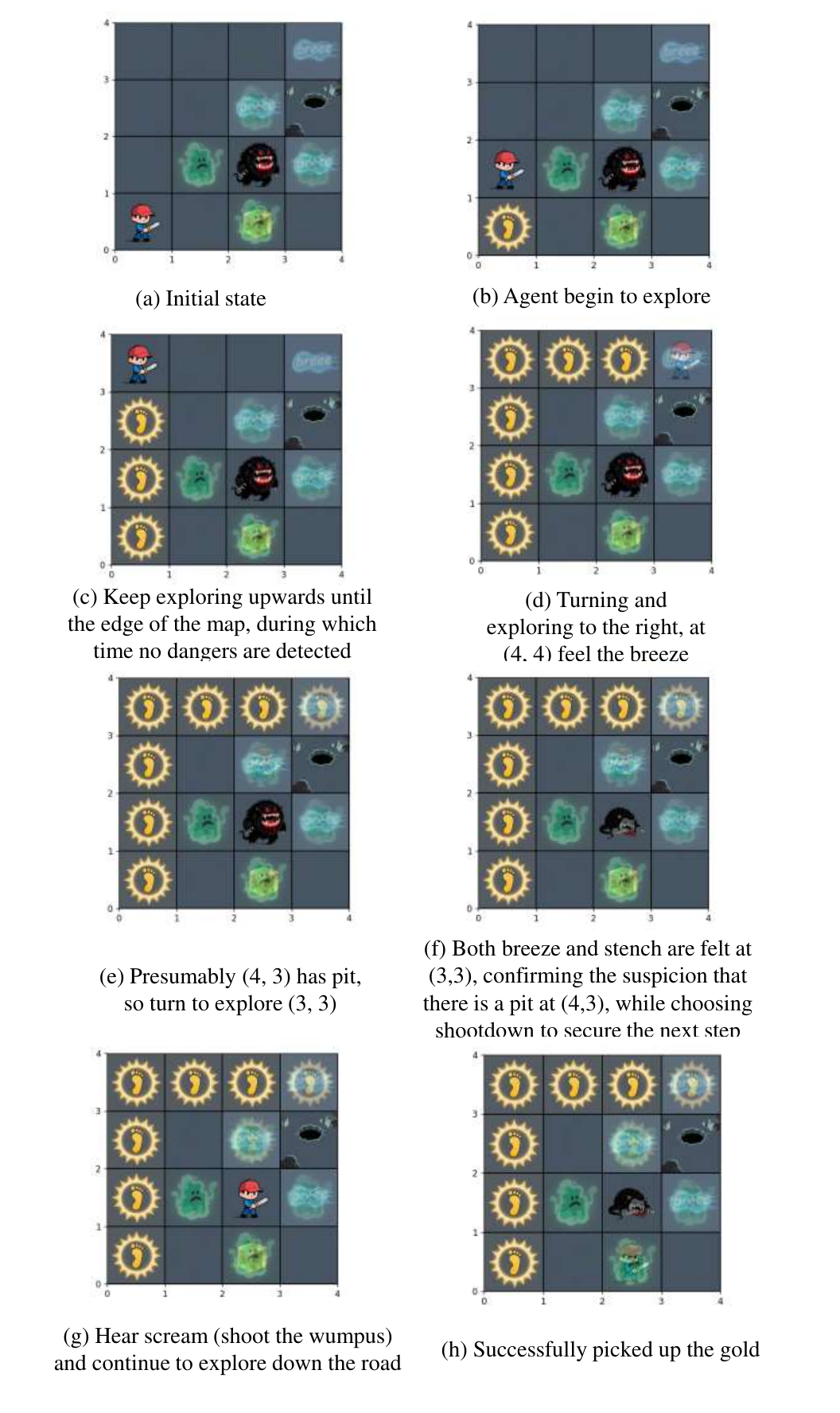}
\caption{\textbf{A typical experiment replay.} 
In LLM-Cave, the DeepSeek R1 model effectively guided the agent through a hazardous cave. Starting from a safe region, the agent detected signs of danger—breeze and stench—and inferred the locations of a pit and the Wumpus. It successfully avoided the pit, eliminated the Wumpus, and proceeded to collect the gold, demonstrating accurate reasoning and decision-making under uncertainty.}
\label{fig4}
\end{figure}

\subsection{Effect of Chain of Speculation on o1-mini}

We first applied the Chain of Speculation method on the o1-mini model to evaluate its impact on performance.

\begin{table}[htbp]
\caption{Effect of Chain of Speculation on o1-mini}
\label{tab:draft_chain_ieee}
\begin{center}
\begin{tabular}{|l|c|c|}
\hline
 & \textbf{o1-mini} & \textbf{o1-mini} \\
\textbf{Metric} & \textbf{ (with only CoT)}& \textbf{ (with CoS\textsuperscript{b})}\\
\hline
Total Runs & 150 & 150 \\
\hline
Avg Total Reward & 72.13 (±35.56) & \textbf{82.16 (±29.02)} \\
\hline
Avg Steps & \textbf{4.87} (range: 1--14) & 5.17 (range: 1--14) \\
\hline
Avg Reward per Step & 23.44 & \textbf{24.34} \\
\hline
Success Rate (\%)\textsuperscript{a} & 65.33\% & \textbf{78.67\%} \\
\hline
Wumpus Kill Rate (\%) & 19.33\% & \textbf{22.00\%} \\
\hline
Avg Latency per Step (s) & \textbf{32.43} & 36.88 \\
\hline
Avg Total Tokens & \textbf{3896.2} & 6631.7 \\
\hline
Avg Cost per Step (USD) & \textbf{0.0015} & 0.0025 \\
\hline
\multicolumn{3}{l}{\textsuperscript{a}Success is defined as the agent successfully collecting the gold.} \\
\multicolumn{3}{l}{\textsuperscript{b}CoS stands for Chain of Speculation.} \\
\end{tabular}
\end{center}
\end{table}

Results indicate that introducing the Chain of Speculation increased the success rate from 65.33\% to 78.67\% and raised the average total reward from 72.13 to 82.16. The performance variance also reduced (standard deviation from ±35.56 to ±29.02), demonstrating more stable behavior. Additionally, the Wumpus kill rate rose from 19.33\% to 22.00\%. Although the average number of steps taken slightly increased (from 4.87 to 5.17), the average reward per step also improved (from 23.44 to 24.34), suggesting better efficiency. 

Regarding computational efficiency, introducing the Chain of Speculation resulted in an increase in computational overhead. Specifically, the average latency per step increased from 32.43s to 36.88s, average token usage per run rose substantially from 3896.2 to 6631.7 tokens, and correspondingly, the average cost per step grew from 0.0015 to 0.0025 USD. This demonstrates a clear tradeoff between improved inference performance and increased computational resource requirements.

\subsection{Application of Planner-Critic on 4o-mini}\label{AA}

Next, we turned to the 4o-mini model. Because the 4o-mini model lacks strong reasoning capabilities, the Chain of Speculation method provided limited benefit. Therefore, we introduced a Planner-Critic framework to further improve this model's performance.

\begin{table}[htbp]
\caption{Effect of Planner-Critic on 4o-mini}
\begin{center}
\begin{tabular}{|l|c|c|}
\hline
 & \textbf{4o-mini} & \textbf{4o-mini} \\
\textbf{Metric} & \textbf{(with only CoT)} & \textbf{(with Critic)} \\
\hline
Total Runs & 150 & 150 \\
\hline
Avg Total Reward & 54.31 (±38.05) & \textbf{60.25 (±37.42)} \\
\hline
Avg Steps & 4.49 (range: 1--17) & \textbf{4.22} (range: 1--13) \\
\hline
Avg Reward per Step & 20.23 & \textbf{23.31} \\
\hline
Success Rate (\%)\textsuperscript{a} & 44.00\% & \textbf{50.67\%} \\
\hline
Wumpus Kill Rate (\%) & \textbf{3.33\%} & 2.00\% \\
\hline
Avg Latency per Step (s) & \textbf{4.79} & 10.14 \\
\hline
Avg Tokens per Run & \textbf{3257.9} & 6256.0 \\
\hline
Avg Cost per Step (USD) & \textbf{0.0029} & 0.0054 \\
\hline
\multicolumn{3}{l}{\textsuperscript{a}Success is defined as the agent successfully collecting the gold.} \\
\end{tabular}
\label{tab:critic_4omini_comparison}
\end{center}
\end{table}

The Planner-Critic approach yields consistent performance gains for 4o-mini: the success rate increased from 44.00\% to 50.67\%, the average total reward rose from 54.31 to 60.25, and the average reward per step improved from 20.23 to 23.31 (approximately a 15\% gain). Notably, introducing a Critic also slightly reduced the average steps taken (from 4.49 to 4.22), indicating the model became more efficient in achieving its goals. Although the Wumpus kill rate dropped from 3.33\% to 2.00\%, the overall increase in success rate suggests a net improvement in performance. 

In terms of computational efficiency, incorporating the Critic agent resulted in higher resource consumption. Specifically, the average latency per step more than doubled (from 4.79s to 10.14s), token usage per run nearly doubled (from 3257.9 to 6256.0 tokens), and average cost per step increased correspondingly (from 0.0029 to 0.0054 USD). These observations demonstrate a clear performance-resource trade-off when integrating the Critic into the planning process.

\subsection{Baseline Performance of Mainstream Models}\label{BB}

Finally, we evaluated the baseline performance of several mainstream large models with no special mechanisms (no active guessing, no Critic) to understand each model's capability under default conditions. Table 3 lists the average reward, success rate, and Wumpus kill rate of five models in this base setting.

\begin{table}[htbp]
\caption{Performance Comparison of Different LLMs in the Wumpus Environment}
\begin{center}
\begin{tabular}{|l|c|c|c|}
\hline
\textbf{Model} & \textbf{Avg Reward} & \textbf{Success Rate \textsuperscript{a}} & \textbf{Wumpus Kill} \\
 &  & \textbf{(\%)} & \textbf{Rate(\%)} \\
\hline
4o-mini         & 54.31 (±38.05)        & 44.00     & 3.33 \\
\hline
o1-mini         & 72.13 (±35.56)        & 65.33     & 19.33 \\
\hline
Gemini 2.5 Flash & 74.73 (±34.30)       & 76.67     & 27.33 \\
\hline
Claude 3.5      & 80.37 (±30.92)        & \textbf{78.00}     & 28.67 \\
\hline
Deepseek R1     & \textbf{81.31 (±26.56)} & 74.00     & \textbf{28.67} \\
\hline
\multicolumn{4}{l}{\textsuperscript{a}Success is defined as the agent successfully collecting the gold.}
\end{tabular}
\label{tab:llm_comparison_short}
\end{center}
\end{table}

As shown in Table 3, the more powerful models exhibit superior and more stable performance across all metrics. For example, Claude 3.5 and Deepseek R1 achieve average rewards above 80, with success rates around 78\% and 74\%, respectively, and Wumpus kill rates of approximately 29\%, significantly outperforming the other models. Gemini 2.5 Flash and o1-mini attain intermediate performance, with success rates of 76.67\% and 65.33\%. In contrast, the smaller 4o-mini model only achieves a 44.00\% success rate and a 3.33\% kill rate, and its score varies widely between runs (standard deviation ±38.05, much higher than those of the stronger models). Overall, these findings indicate that the score of this environment can effectively reflect the reasoning ability of the model. 

\begin{table}[htbp]
\caption{Computation Efficiency Metrics of Different LLMs in the Wumpus Environment}
\begin{center}
\begin{tabular}{|l|c|c|c|c|c|}
\hline
\textbf{\shortstack{Metric}} 
& \shortstack{\textbf{4o-}\\\textbf{mini}} 
& \shortstack{\textbf{Claude}\\\textbf{3.5}} 
& \shortstack{\textbf{Deepseek}\\\textbf{R1}} 
& \shortstack{\textbf{Gemini}\\\textbf{2.5 Flash}} 
& \shortstack{\textbf{o1-}\\\textbf{mini}} \\
\hline
\shortstack{Avg\\Latency (s)} 
& 4.79 
& 11.11 
& 49.42 
& 21.52 
& 32.43 \\
\hline
\shortstack{Tokens\\Prompt} 
& 1940.6 
& 2764.9 
& 2144.3 
& 2578.4 
& 2130.1 \\
\hline
\shortstack{Tokens\\Completion} 
& 1317.3 
& 2987.7 
& 14027.0 
& 38513.8 
& 1766.1 \\
\hline
\shortstack{ Tokens\\Total} 
& 3257.9 
& 5752.6 
& 16171.3 
& 41092.2 
& 3896.2 \\
\hline
\shortstack{Avg\\Cost (USD)} 
& 0.0029 
& 0.0531 
& 0.0147 
& 0.0796 
& 0.0015 \\
\hline
\shortstack{TPS\textsuperscript{a}\\(tokens/s)} 
& 47.25 
& 30.97 
& 36.82 
& 195.51 
& 8.16 \\
\hline
\multicolumn{6}{l}{\textsuperscript{a}TPS = Tokens per second.}
\end{tabular}
\label{tab:llm_efficiency_transposed}
\end{center}
\end{table}

In addition to task performance, computation efficiency metrics (Table 4) provide further insights into practical trade-offs among these models. Claude 3.5, despite its strong performance, incurs significantly higher computational costs (\$0.0531 per run) compared to other models, primarily due to a larger average token usage (5752.6 tokens per run). In contrast, models such as 4o-mini and o1-mini, although exhibiting lower success rates, are markedly more economical (\$0.0029 and \$0.0015 per run, respectively), reflecting their smaller size and lower computational overhead. Gemini 2.5 Flash demonstrates exceptionally high decoding speed, averaging 195.51 tokens per second (TPS), which significantly surpasses all other tested models, highlighting its suitability for time-sensitive or interactive applications. Conversely, o1-mini shows a notably low TPS of 8.16, suggesting limited practical deployment in latency-critical scenarios despite acceptable overall reasoning ability. Deepseek R1 balances performance and cost relatively effectively (\$0.0147 per run), although its average latency per step (49.42s) remains substantially higher than others, indicating potential drawbacks in real-time use cases. Taken together, these computational efficiency metrics complement traditional accuracy metrics and provide another perspective on the performance of large models.

\section{Discussion}

Our experiments demonstrate that the LLM-Cave environment is effective in evaluating the multi-step reasoning capabilities of large language models. The effectiveness of the reasoning enhancement strategy depends on the underlying reasoning strength of the model. For example, o1-mini benefits greatly from “Chain of Speculation” (CoS), with an increase in success rate of about 13\% and an increase in average score of 10 points. This suggests that the CoS mechanism can leverage explicit reasoning structures to optimize decision making.

In contrast, the weaker 4o-mini model gains less from CoS due to the inaccuracy of the initial guess propagation. Instead, it benefits more significantly from the Planner-Critic mechanism, which improves the success rate by about 6.7\% and the average score by 6 points by directly optimizing decisions and correcting errors. Thus, stronger models can maximize their potential through CoS, while weaker models can gain more from “planner-critic”, although they are still constrained by their inherent limitations.

In terms of computational efficiency, both strategies incur additional resource overheads, including increased latency and token usage, which is particularly noticeable in the weaker models employing Planner-Critic. However, the substantial performance gains justify these incremental costs, making these strategies a worthwhile investment for tasks requiring higher accuracy and optimized decision making.

\section{Conclusion}

We have developed LLM-Cave, a novel benchmark framework specifically designed to evaluate and enhance the reasoning and decision-making capabilities of Large Language Models (LLMs). In our experiments, we introduced complementary inference, the Chain of Speculation prompting technique and an LLM-based Planner-Critic feedback framework. Experimental results indicate that explicitly integrating structured reasoning processes, whether through chain-of-thought prompting or evaluative feedback mechanisms, significantly helps LLMs achieve more accurate, reward-maximizing outcomes in multi-step reasoning scenarios. The computational cost associated with these strategies, while noticeable, remains a worthwhile trade-off given the observed performance gains. Looking forward, integrating inference-time reasoning strategies (such as Chain of Speculation and Planner-Critic) with training-time feedback or fine-tuning (e.g., incorporating Critic feedback into reinforcement learning or supervised fine-tuning) offers a promising path toward enhancing autonomy and reliability in complex reasoning tasks.

\end{document}